\documentclass[runningheads]{llncs}
\usepackage[utf8]{inputenc}
\usepackage{graphicx}
\usepackage{amssymb}
\usepackage[font=small,labelfont=bf]{caption}
\usepackage[font=footnotesize]{subfig}
\usepackage{mathtools}
\usepackage{color}
\usepackage{ulem}
\DeclarePairedDelimiterX{\norm}[1]{\lVert}{\rVert}{#1}

\usepackage{hyperref}

\makeatletter
\newcommand{\printfnsymbol}[1]{%
  \textsuperscript{\@fnsymbol{#1}}%
}
\makeatother
\begin{document}
\title{Semi-Supervised Learning by Disentangling and Self-Ensembling over Stochastic Latent Space 
}



\author{Prashnna Kumar Gyawali\printfnsymbol{1}\thanks{\printfnsymbol{1}Both authors contributed equally to this work}
\and Zhiyuan Li\printfnsymbol{1} \and Sandesh Ghimire \and Linwei Wang}
\institute{\email{pkg2182@rit.edu, zl7904@rit.edu} \\ Rochester Institute of Technology}

\renewcommand\footnotemark{}
%
%
\maketitle              
\begin{abstract}
The success of deep learning in medical imaging is mostly achieved at the cost of a large labeled data set.
Semi-supervised learning (SSL) provides a promising solution by leveraging the structure of unlabeled data to improve learning from a small set of labeled data.
Self-ensembling 
is a simple approach used in SSL to encourage consensus among ensemble predictions of unknown labels, 
improving generalization of the model by making it more insensitive to the latent space. 
Currently, such ensemble is obtained by randomization such as dropout regularization and random data augmentation. 
In this work, we hypothesize -- from the generalization perspective -- that self-ensembling can be improved by exploiting the stochasticity of a disentangled latent space. 
To this end, we 
present a stacked SSL model that 
utilizes unsupervised disentangled representation learning as the stochastic embedding for self-ensembling. 
We evaluate the presented model for multi-label classification using chest X-ray images, demonstrating its improved performance over related SSL models 
as well as the interpretability of its disentangled representations. 

\end{abstract}

\section{Introduction}
While deep learning has seen tremendous successes in a variety of medical image analysis problems \cite{litjens2017survey}, many of these successes are achieved at the cost of a large labeled data set. 
Labeling a large collection of data  however requires  
substantial human and financial resources, creating a primary hurdle to the wide-spread adoption of deep learning in clinical practice \cite{litjens2017survey}. Semi-supervised learning (SSL) provides a promising solution 
by leveraging the structure of unlabeled data to improve the  learning from a small set of labeled data \cite{kingma2014semi,laine2016temporal}.

Among existing SSL methods, 
self-ensembling 
is a simple approach that encourages 
consensus 
among ensemble predictions 
of unknown labels \cite{laine2016temporal}.
Such ensemble predictions can be formed by randomization such as 
network regularization (\textit{e.g.}, dropout) 
and random input augmentation  \cite{laine2016temporal}. 
As later rationalized in \cite{kawaguchi2018towards}, 
based on the analytical learning theory,
these randomization techniques are critical 
as 
they improve the generalization of the model by making it more \textit{insensitive} to the latent space.
From this theoretical perspective, 
it is natural to hypothesize that 
the design of this randomization will benefit from the knowledge of the latent space, 
especially its stochasticity. 
This however is not considered in 
existing works.  
The input augmentation functions, 
for instance, 
are typically hand-crafted 
considering
random translations or rotations of images 
\cite{laine2016temporal}, with little 
consideration to the distribution of latent variables.

In parallel, advances in representation learning -- especially that of the variational auto-encoder (VAE) -- has allowed us to 
infer posterior distributions of the latent variables
in an unsupervised manner \cite{kingma2013auto}. In the classic semi-supervised deep generative model presented in \cite{kingma2014semi}, 
it has also been shown that 
such an unsupervised embedding 
can largely 
facilitate the subsequent SSL training 
by providing a 
disentangled and thereby more separable latent space 
\cite{kingma2014semi}.



In this paper, 
drawing on the analytical  learning theory \cite{kawaguchi2018towards}, we rationalize that 
1) disentangling and 2) self-ensembling over the stochastic latent space 
will improve the generalization ability of the model. 
Based on this rationale,
we investigate using unsupervised disentangled representation learning 
as the stochastic input embedding in 
self-ensembling. 
The presented SSL model 
consists of a VAE-based unsupervised embedding of the data, 
followed by a semi-supervised 
self-ensembling network utilizing 
the stochastic embedding as the inherent random augmentation of the inputs. 
We evaluate the presented SSL model 
on the recently open-sourced Chexpert data set for \textit{multi-label} classification of thoracic disease using chest X-rays \cite{irvin2019chexpert}. 
To demonstrate the benefits gained by exploiting the stochastic latent space in self-ensembling, 
we compare the performance of the presented method with the standard self-ensembling method considering different image-level input augmentation methods \cite{laine2016temporal}, VAE-based embedding with and without a subsequent deep generative SSL \cite{kingma2014semi}, along with a generative adversarial network (GAN)-based SSL \cite{odena2017conditional}.
We further qualitatively demonstrate the disentangled representation obtained via unsupervised embedding, and discuss its use for data analysis and model interpretability. 

\section{Related Work} 

This work is mostly related to two lines of research: 1) SSL based on regularization with random transformations, and 2) disentangled representation learning and its use in SSL. 
In the former, 
consistency-based regularization is applied on ensemble predictions obtained 
by randomization 
techniques 
such as random data augmentation, dropout, and random max-pooling 
\cite{laine2016temporal}.
This randomization was empirically shown to improve the generalization and stability of the SSL model, 
while its theoretical basis was recently shown to be related to the reduction of model sensitivity to the latent space \cite{kawaguchi2018towards,ghimire2019improving}. 
Motivated by this theory, 
in this work, 
we attempt to 
utilize the knowledge about the stochastic latent space -- obtained in unsupervised learning -- 
in this randomization process. 

In the latter  representation learning, 
deep neural networks have been combined with
variational inference 
to jointly realize generative modeling of unlabeled data and posterior inference of 
latent variables \cite{kingma2013auto}. 
Furthermore, 
the learned latent representations are encouraged to be semantically interpretable and mutually invariant, which is empirically shown
to be useful for the downstream tasks \cite{kingma2014semi}.
For instance, an unsupervised VAE was used to provide a disentangled embedding (M1) for a subsequent VAE-based semi-supervised model (M2), 
commonly known as the M1+M2 model \cite{kingma2014semi}. 
In this work,
we make the first attempt to use the analytical learning theory to support the effect a disentangled embedding can have on the generalization ability of a model. 
Furthermore, we improve
the M1+M2 model by replacing M2 with a self-ensembling SSL network, 
taking VAE's ability to model stochastic latent space 
to support self-ensembling. 


Besideds the approaches discussed above,
there is also an active line of research in GAN-based SSL methods \cite{odena2017conditional,madani2018semi}.
The general idea is to add a classification objective to the original mini-max game and increases the capacity of the discriminator to associate the inputs to the corresponding labels. 
The presented work differs from this line of research by the emphasis on obtaining, 
regularizing, 
and interpreting the latent representations in SSL.


An increased interest in SSL has also been seen in medical image anlaysis.  
The use of an unsupervised representation learning for better generalization has been investigated for the task of myocardial segmentation \cite{chartsias2018factorised}. 
In \cite{madani2018semi}, SSL was used in a similar X-ray data set, although the scope was limited to binary classifications between normal and abnormal categories. 
To our knowledge, we present the first multi-label SSL that investigates disentangled learning and self-ensembling of stochastic latent space in medical image classification.


\section{Model}

\begin{figure}[b]
\centering
    \includegraphics[width=0.91\textwidth]{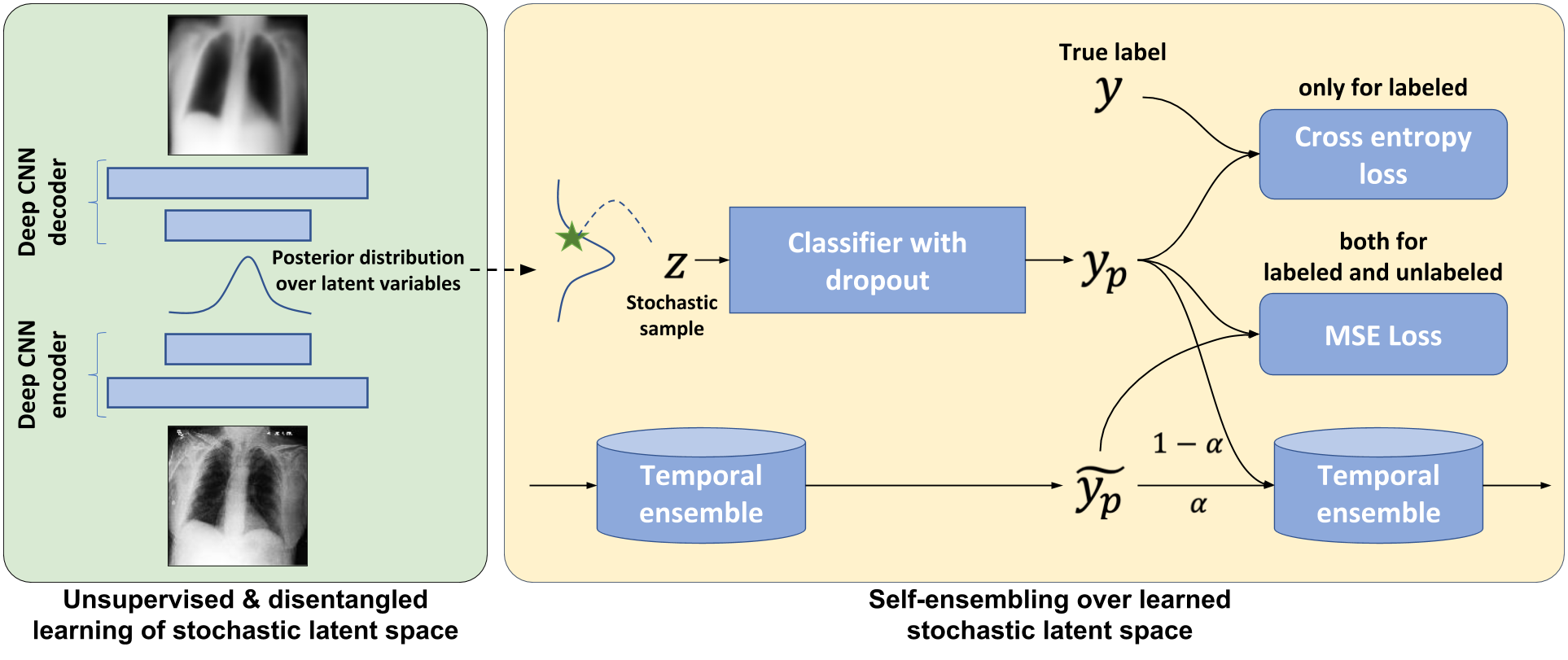}
    \caption{Schematic diagram of the presented model.
    }
    \label{fig:model}
\end{figure}

We consider training data $\mathcal{D}$ = $\mathcal{D}_{l} \cup \mathcal{D}_{u}$, where $\mathcal{D}_l$ = $\{\mathbf{x}_i ,\mathbf{y}_i\}^{N_{l}}_{i=1}$ is the labeled set and $\mathcal{D}_u$ = $\{\mathbf{x}_j\}^{N_{u}}_{j=1}$ the unlabeled set. As outlined in Fig.~\ref{fig:model}, a stochastic latent space will be learned in an unsupervised and disentangled manner (section \ref{sec:unsupervised}), 
which will then be regularized via self-ensembling for SSL (section \ref{sec:ensemble}).

\subsection{\small{Unsupervised \& Disentangled
Learning of Stochastic Latent Space}}
\label{sec:unsupervised}
To disentangle and obtain the posterior distribution in the latent space, we first utilize a VAE for unsupervised learning of the generative factors in the data. We assume data $\mathbf{x}$ to be generated by a likelihood $p_{\mathbf{\theta}}(\mathbf{x}|\mathbf{z})$, involving a latent variable $\mathbf{z}$ with a certain prior distribution $p(\mathbf{z})$.
Given the intractability of exact posterior inference, 
a distribution $q_{\phi}\mathbf{(z|x)}$ is introduced to approximate the true posterior  $p\mathbf{(z|x)}$ using variational inference \cite{kingma2013auto}. Both the 
approximated posterior 
$q_{\phi}\mathbf{(z|x)}$ 
and the likelihood 
$p_{\theta}(\mathbf{x}|\mathbf{z})$
 are parameterized by deep neural networks.

The VAE is trained by maximizing the  variational evidence lower bound  of the marginal likelihood of the training data
with respect to parameters $\theta$ and $\phi$:
\begin{equation}
  \label{eq:VAEObj}
  \log{p}(\mathbf{x}) \geq \mathcal{L}  = \mathbb{E}_{q_{\phi}(\mathbf{z}|\mathbf{x})}[\log{p_{\mathbf{\theta}}(\mathbf{x}|\mathbf{z})}] - KL(q_{\phi}(\mathbf{z}|\mathbf{x})||p(\mathbf{z})) 
\end{equation} 
where the first term can be interpreted as minimizing the reconstruction error, and the second term 
regularizes the learned posterior density $q_{\phi}(\mathbf{z}|\mathbf{x})$ 
by a prior $p(\mathbf{z})$ 
via the Kullback-Leibler (KL) divergence measure. 
We set the prior $p(\mathbf{z})$ to be an isotropic Gaussian which, 
through mutually independent latent dimensions, encourages disentangled latent representations in $q_{\phi}\mathbf{(z|x)}$  \cite{kingma2013auto}.

\subsection{\small{Regularizing the Stochastic Latent Space via Self-Ensemble SSL}}
\label{sec:ensemble}
The unsupervised VAE embedding is then used as the input to a self-ensembling SSL model which, 
on unsupervised targets, 
applies coherence-based regularization to form consensus among ensemble predictions. 
For each training sample $\mathbf{x}^{(i)}$, 
its ensemble predictions are obtained from three sources: 
\begin{enumerate}
\item 
Sampling from VAE-learned posterior density $q(\mathbf{z}^{(i)}|\mathbf{x}^{(i)})$. 
 This utilizes a distribution learned from unlabeled data 
 to replace the commonly-used  hand-crafted augmentation functions to perturb $\mathbf{x}^{(i)}$ \cite{laine2016temporal}.
\item Network dropout that randomly neglects some units and utilizes a sub-network during training. 
\item 
Temporal ensemble \cite{laine2016temporal} achieved by accumulating the predicted label $\mathbf{y}_{p}$ after every training epoch into an ensemble output $\tilde{\mathbf{y}}_{p}$ by $\tilde{\mathbf{y}}_{p} \leftarrow \alpha \tilde{\mathbf{y}}_{p} + (1 - \alpha)\mathbf{y}_{p}$, where $\alpha$ 
controls how far the ensemble reaches into training history.
\end{enumerate}

Given each pair of ensemble predictions $\mathbf{y}_{p}$ and $\mathbf{\tilde{y}}_{p}$, the network is trained, in each batch $B$, with the objective of minimizing the ensemble loss $\mathcal{L}^{e}$:
\[
\mathcal{L}^{\text{e}} = \underbrace{\frac{1}{|B|}\sum_{n\sim(B\cap \mathcal{D}_{l})}
\sum^{L}_{l=1}\big[ - y_{n,l}\log{f(y_{n,p}|q(\mathbf{z}|\mathbf{x}))}\big]}_{\text{only for labeled}}  + \underbrace{\zeta \cdot \frac{1}{|B|}\sum_{n\sim B} \norm[\big]{\mathbf{y}_{p} - \tilde{\mathbf{y}}_{p}}^2}_{\text{both for labeled and unlabeled}}  (2)
\]
where the first term corresponds to the standard cross entropy loss and is evaluated 
for labeled data. The second term is evaluated for all data,  encouraging consensus among ensemble predictions via mean squared loss. 
A ramp-up weighted function, starting from zero, is used for $\zeta$ as described in \cite{laine2016temporal}.

\subsection{\small{Stacked Self-Ensembling SSL from the Generalization Perspective}}

In this section, we examine the generalization ability of the presented method -- via the recently introduced analytical learning theory \cite{kawaguchi2018towards} -- from two viewpoints: self-ensembling and disentangling. 
Theorem 1 in \cite{kawaguchi2018towards} provides an upper bound on the generalization gap (difference between expected and empirical error) $\Delta_{g}$:
\[
\Delta_{g} \leq V[f] \cdot \mathbb{D}^{*} \tag{3}
\]
where $V[f]$ is the \textit{variation} that computes how a function $f$ varies in total w.r.t each small perturbation of every cross-combination of its variables, and $\mathbb{D}^{*}$ is the \textit{discrepancy} between the latent projections of an available data set $\mathcal{D}$ and true data distribution. 
In our case, $f$ is the composition of the loss function $\ell$ of coherence-based regularization and the mapping function $f_{y}$ between stochastic latent sample $\mathbf{z}$ and the prediction $\mathbf{y}_p$, \textit{i.e.,} $f = \ell \circ f_{y}$. 
Here, we rationalize how the presented method decreases the generalization gap by decreasing $V[f_y]$.

\subsubsection{Self-ensembling:} The second regularization term in
(2)
can be re-written over the input samples drawn from the posterior density $q(\mathbf{z}|\mathbf{x})$ from VAE embedding:
\[
\ell = \int_{\mathbf{z_1}, \mathbf{z_2}} \norm[\big]{f_y(\mathbf{z_1}) - f_y(\mathbf{z_2})}^2 dP(\mathbf{z_1}, \mathbf{z_2}|\mathbf{x})
\]
where $P(\mathbf{z_1}, \mathbf{z_2}|\mathbf{x})$ = $q(\mathbf{z_1}|\mathbf{x})q(\mathbf{z_2}|\mathbf{x})$. 
Minimizing this loss explicitly requires  
$f_y$ to be more insensitive over the space of $\mathbf{z}$ 
which 
implicitly minimizes $V[f_y]$ and thus the bound on the generalization gap.

\subsubsection{Disentangling:} 
Given representation $(\mathbf{z}_y, \mathbf{z}_o)$ where $\mathbf{z}_y$ and $\mathbf{z}_o$ are respectively the latent variables related and unrelated to $\mathbf{y}$. 
Based on \cite{kawaguchi2018towards}, the \textit{variation} $V[f_y]$ is minimal when the mapping $f_{y}$ from latent space $(\mathbf{z}_y, \mathbf{z}_o)$ is invariant over 
$\mathbf{z}_o$. 
A disentangled latent space by design thereby reduces the generalization gap.

\section{Experiments}
\subsubsection{Data and Model Architecture:}
We evaluated the presented model on the recently open-sourced Chexpert data set that has strong reference standards compared to other similar large-scale chest X-ray data set \cite{irvin2019chexpert}.
It consists of 224,316 chest radiographs from 65,240 patients, with labels for 14 pathology categories extracted from radiology reports. 
Given uncertainty labels provided for all images, we first removed all uncertain samples from the data set. We also removed all lateral-view samples.
Small labeled training sets were then created by balancing among each disease category, ranging from 100 to 500 samples per category.
Another   
5000 and $5000 \times 10$ samples were randomly selected as the validation and test sets, 
while the rest were used as unlabeled training data. 
All images were re-sized to $128 \times 128$ in dimension.

The encoder of the VAE had five convolutional layers followed by three fully-connected (FC) layers. 
The output of the last FC layer was divided into the mean and log variance of the posterior distribution of the latent variables. 
The decoder is symmetrical to the encoder, with two FC layers followed by five transposed convolutional layers. 
ReLU activation was used for all the layers, except the last encoder layer that has no activation and the last decoder layer that used sigmoid activation.
The self-ensembling network 
consisted of three FC layers 
with dropout layers ($p=0.5$) in between. 
ReLU activation was used for the first two FC layers, and sigmoid activation for the last one. The schematic diagram of the architecture is provided in the Appendix (Fig. \ref{fig:archi}).
We use the Adam optimizer throughout all our experiments with a learning rate of $1e$-5. 

\subsection{Multi-Label Semi-Supervised Classification}

\begin{table}[t]
\label{classification_table}
\centering
    \begin{tabular}[t]{|l|c|c|c|c|c|}
    \hline
    Model & \multicolumn{5}{c|}{Size of labeled data set ($k$)} \\ \cline{2-6}
    & 100 & 200 & 300 & 400 & 500\\
    \hline
    VAE embedding        & 0.5853 & 0.6119 & 0.6331 & 0.6416 & 0.6556 \\
    VAE embedding + SSL (M1 + M2) & 0.6080 & 0.6272 & 0.6340 & 0.6390 & 0.6491 \\
    Image-space self-ensemble (with noise) & 0.6012  & 0.6277  & 0.6444  & 0.6550 & 0.6626 \\
    Image-space self-ensemble (with augmentation)  & 0.6089  & 0.6301  & 0.6423  & 0.6530 & 0.6617 \\
    ACGAN   & 0.5865   &0.6036   &0.6064   &0.6284 & 0.6372   \\
    \textbf{Latent-space self-ensemble (presented)}   & \textbf{0.6200} & \textbf{0.6386} & \textbf{0.6484} & \textbf{0.6637} & \textbf{0.6697} \\
    \hline
    \end{tabular} 
    \caption{Mean AUROC for multi-label classification for 14 pathology categories. trained with a fixed number of unlabeled data and a varying number of labeled data.
    }
\end{table}
We quantitatively evaluated the SSL performance of the presented model in comparison to existing models as summarized in table 1.
We first trained a classifier (with the same architecture as the self-ensembling network) using the 
unsupervised VAE embedding on labeled training data (\textit{VAE embedding}).
We then considered two most-related models to the presented work: 
1) the M1+M2 SSL model where the unsupervised VAE embedding was used to support a subsequent VAE-based SSL model (\textit{VAE embedding + SSL}) \cite{kingma2014semi}, 
and 2) the self-ensembling SSL
(with the same architecture as the VAE encoder followed by self-ensemling in the stacked model) 
with two different types of image-level augmentation: 
adding Gaussian noises (with $std = 0.15$) or random translation and rotation (maximum 12 pixels and 10 degrees) to the images (\textit{image-space self-ensembling}) \cite{laine2016temporal}. 
To demonstrate how the presented work relates to GAN-based SSL methods, we also 
added a comparison to 
ACGAN  \cite{odena2017conditional} that was shown to generate globally coherent and discriminative samples assisting in SSL. 

\begin{figure}[t]
\centering
    \includegraphics[width=0.8\textwidth]{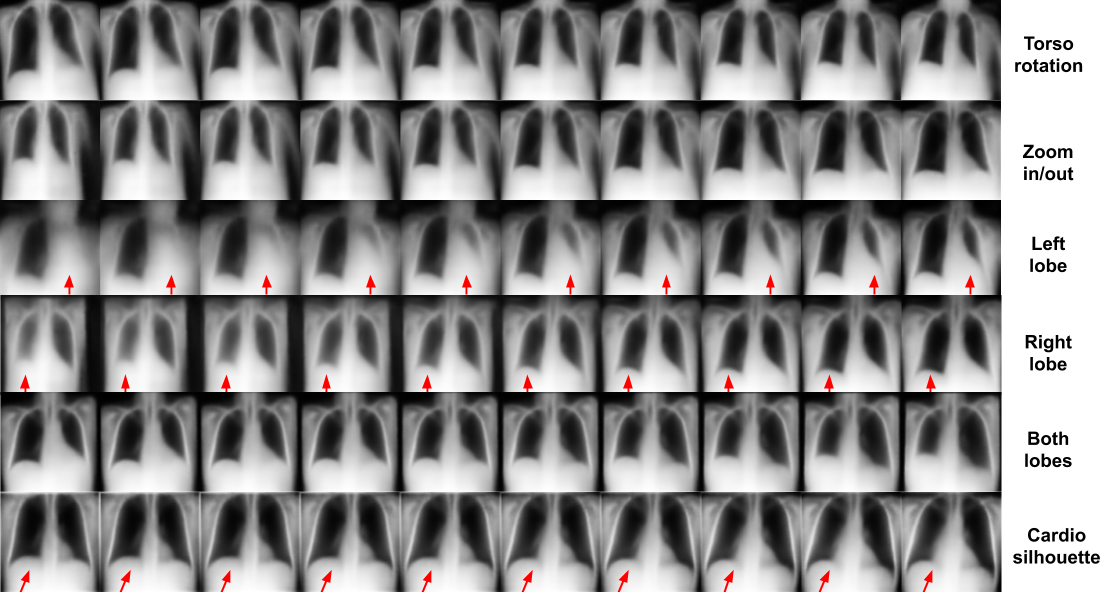}
    \caption{Images generated by traversing over the [-3, 3] range of a specific latent dimension, based on latent values inferred from a different seed image for each row.}
    \label{fig:reconstruction}
\end{figure}

We tested all models on a varying number of labeled training data whilst keeping the number of unlabeled data fixed. Each model was tested on 10 different testing sets (5000 samples each).
As shown, the three self-ensembling methods in general achieved better performance. 
Among the three self-ensembling methods, all the standard deviation is less than 0.008, and the improvement of the presented method over the two baseline self-ensembling methods is statistical significant ($p<0.05$).
This verified our hypothesis that, in comparison to ad-hoc image-level augmentations, 
utilizing the stochasticity of the latent space can improve the performance of self-ensembling. 

It was surprising that ACGAN performed worse (except $k = 100$) than VAE embedding. We speculated that, 
unlike natural images where ACGAN has seen superior performance, disease-related factors in X-ray images may be more difficult to capture 
among other disease-irrelevant factors 
(see section 5). 

\begin{figure}[t]
\centering
    \includegraphics[width=0.8\textwidth]{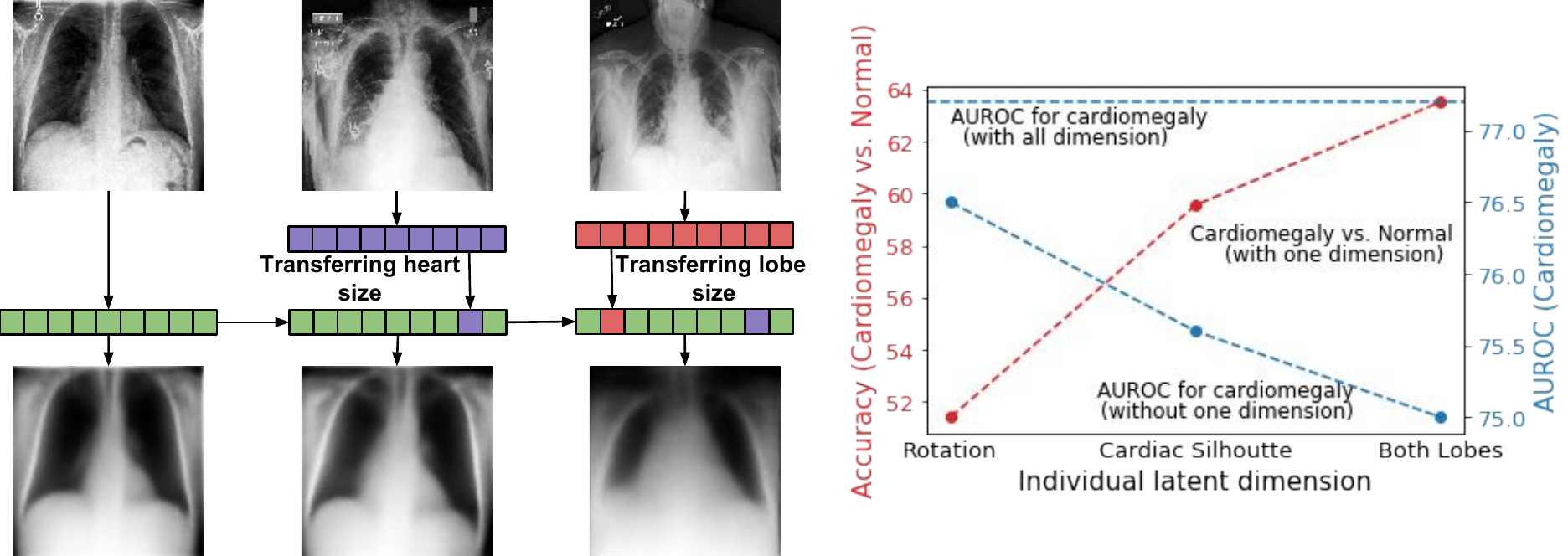}
    \caption{[Best viewed in color] Left: Demonstration of feature transfer by exchanging specific latent units. Top row: original images. Bottom row:   reconstructed with original  or  exchanged latent codes. Right: Discriminative power of individual latent dimensions.  }
    \label{fig:feature_transfer}
\end{figure}

\subsection{Interpreting the Disentangled Latent Representations}

We then qualitatively evaluated the stochastic latent variables learned from the VAE.
As shown in Fig.~\ref{fig:reconstruction}, 
as we traversed along a particular latent dimension 
and kept the others fixed, 
we were able to generate images reflecting changes in a particular semantic factor.
With this knowledge, 
in Figure \ref{fig:feature_transfer} (left),  we demonstrated 
that it is 
possible to 
transfer specific features 
(such as heart size and lobe size)
between X-ray images 
by swapping the corresponding latent units.


In an attempt to quantify 
how these unsupervised disentangled representations may affect downstream classification tasks, 
we considered the pathology of 
cardiomegaly as an example which is reflected as changes in heart-chest ratio in X-ray images \cite{battler1980initial}. 
We built a binary classifier between the category of cardiomegaly and no-finding,  \textit{considering only one of the learned latent units at a time}. 
We randomly sampled respectively 500, 1000 and 2000 images that had either cardiomegaly or no-finding labels for training, validation, and testing. 
As shown in Fig.~\ref{fig:feature_transfer}(right: red curve), 
the more a specific unit  captured the heart-chest ratio, the more discriminative it was for detecting cardiomegaly.
We also re-trained the presented model (for k =500) by removing one latent unit at a time, and evaluated the resulting AUROC for cardiomegaly. 
Similarly (Fig.~\ref{fig:feature_transfer}(right: blue curve)), 
the more a specific unit  captured the heart-chest ratio,
the larger its removal caused the drop in AUROC.
These results suggest that 
improved disentangling in the latent representations 
may facilitate down-stream classification tasks as well as increase the interpretability of the results. 


\section{Conclusions and Discussion}
We presented a stacked SSL method that uses unsupervised disentangling of the stochastic latent space as the input randomization in self-ensembling. 
From the analytical learning theory, we rationalized the effect of disentangling and self-ensembling over the latent space on 
the generalizability of the model. 
Empirically, we demonstrated 
both the quantitative improvement of the presented model in SSL 
and the interpretability of its disentangled representations. 

We noted that, 
compared to many visual benchmark data sets,  
disease-specific factors in medical images 
may be buried by other more significant factors of variations 
in terms of contribution to pixel reconstruction 
or image distribution 
(\textit{e.g.}, heart-chest ratio \textit{vs}.\ torso shape). 
For instance, 
we attempted to remove the inactive dimensions (defined as $A_u < 10^{-2}$ where $A_u$ = $Cov_{x}(\mathbb{E}_{u \sim q(u|x)}[u])$ for each dimension $u$ in $\mathbf{z}$) from VAE embedding, 
a strategy shown to improve the performance of the M1+M2 model in visual benchmarks \cite{kingma2014semi}.
The mean AUROC of the presented model, however, decreased around 3\% to 0.658 (for $k$=500). 
This, we believe, may explain the relatively limited progress of unsupervised representation learning in medical images despite its recent traction in other visual domains, a pressing challenge to be resolved 
in order to leverage unlabeled data in a field where image labeling is especially costly and difficult. For future work, we plan to improve two-stage training strategy and disentangling by hierarchical generative models.



\noindent\textbf{Acknowledgement.} This work is supported by NSF CAREER ACI-1350374 and NIH NHLBI R15HL140500

\bibliographystyle{splncs04}
\bibliography{refs}
\clearpage
\section{Appendix}
\begin{figure}[]
\centering
    \includegraphics[width=1.0\textwidth]{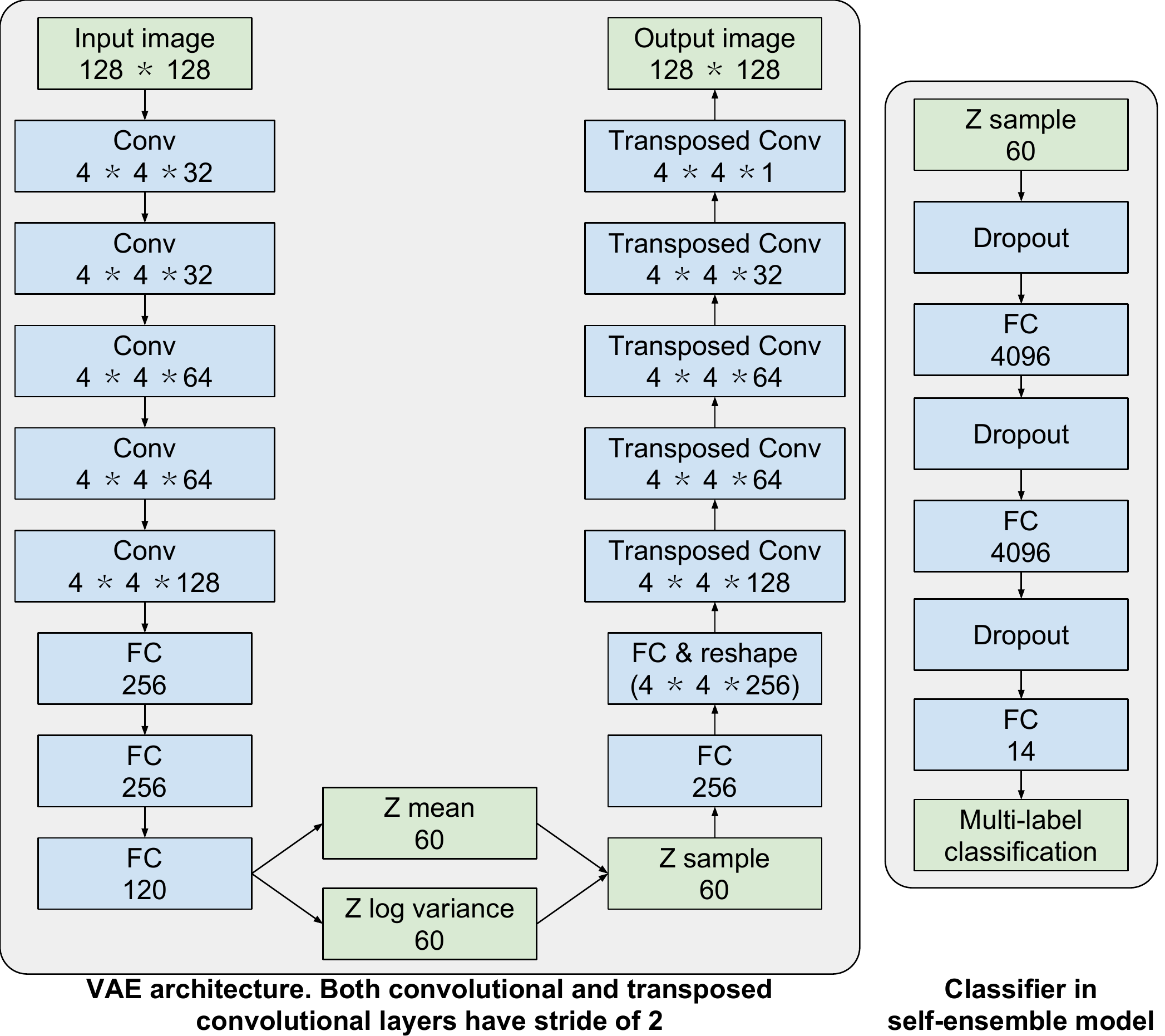} 
    \caption{Schematic diagram of the presented model.
    }
    \label{fig:archi}
\end{figure}
\end{document}